\documentclass[authoryear,times,review,10pt]{elsarticle}
\usepackage{amssymb}

\usepackage{amsmath}
\usepackage{makecell}
\usepackage{booktabs}       
\usepackage{nicefrac}       
\usepackage{microtype}      
\usepackage{multirow}
\usepackage{tabularx}
\usepackage{adjustbox}
\usepackage[table,xcdraw]{xcolor}
\usepackage{tabularx}
\usepackage[hyphens]{url}
\usepackage{hyperref}
\usepackage{algorithm}
\usepackage{algorithmic}
\usepackage{breqn}

\usepackage{graphicx} 
\usepackage{multirow} 
\usepackage{booktabs} 
\newcommand{\invisiblefootnote}[1]{%
    \begingroup%
    \renewcommand\thefootnote{\color{white}1}%
    \footnotetext{#1}%
    \endgroup%
}

\journal{Pattern Recognition}
\begin{document}

\begin{frontmatter}


\title{Boosting Illuminant Estimation in Deep Color Constancy through Enhancing Brightness Robustness}



\author[label1]{Mengda Xie} 
\author[label1]{Chengzhi Zhong} 
\author[label2,label3]{Yiling He} 
\author[label2]{Zhan Qin} 
\author[label1]{Meie Fang}

\affiliation[label1]{organization={School of Computer Science and Cyber Engineering, Guangzhou University},
            city={Guangzhou},
            postcode={510006}, 
            state={Guangdong},
            country={China}}
\affiliation[label2]{organization={State Key Laboratory of Blockchain and Data Security, Zhejiang University},
            city={Hangzhou},
            postcode={310027},
            state={Zhejiang},
            country={China}}
\affiliation[label3]{organization={Information Security Research Group, University College London},
            city={London},
            postcode={NW1 2AE},
            country={United Kingdom}}


\begin{abstract}
\invisiblefootnote{Corresponding author: Meie Fang (fme@gzhu.edu.cn)}
Color constancy estimates illuminant chromaticity to correct color-biased images. Recently, Deep Neural Network-driven Color Constancy (DNNCC) models have made substantial advancements.
Nevertheless, the potential risks in DNNCC due to the vulnerability of deep neural networks have not yet been explored.
In this paper, we conduct the first investigation into the impact of a key factor in color constancy—brightness—on DNNCC from a robustness perspective.
Our evaluation reveals that several mainstream DNNCC models exhibit high sensitivity to brightness despite their focus on chromaticity estimation.
This sheds light on a potential limitation of existing DNNCC models: their sensitivity to brightness may hinder performance given the widespread brightness variations in real-world datasets.
From the insights of our analysis, we propose a simple yet effective brightness robustness enhancement strategy for DNNCC models, termed BRE. The core of BRE is built upon the adaptive step-size adversarial brightness augmentation technique, which identifies high-risk brightness variation and generates augmented images via explicit brightness adjustment. Subsequently, BRE develops a brightness-robustness-aware model optimization strategy that integrates adversarial brightness training and brightness contrastive loss, significantly bolstering the brightness robustness of DNNCC models.
BRE is hyperparameter-free and can be integrated into existing DNNCC models, without incurring additional overhead during the testing phase.
Experiments on two public color constancy datasets—ColorChecker and Cube+—demonstrate that the proposed BRE consistently 
enhances the illuminant estimation performance of existing DNNCC models, reducing the estimation error by an average of 5.04\% across six mainstream DNNCC models, underscoring the critical role of enhancing brightness robustness in these models.

\end{abstract}


\begin{highlights}
\item We first propose the concept of “brightness vulnerability” for color constancy. By systematically evaluating the DNNCC model's robustness to brightness variations, we identified it as a potential cause of model degradation.
\item We propose BRE, an incremental plug-in designed to enhance the brightness robustness of existing DNNCC models. BRE adaptively generates adversarially augmented brightness images via parameterized brightness curves, and further integrates brightness-robustness-aware optimization to enhance the performance of DNNCC models under diverse brightness variations.
\item Comprehensive evaluations on two real-world color constancy datasets indicate that the proposed BRE consistently enhances the performance of mainstream DNNCC models, without incurring additional computational overhead during the testing phase.
\end{highlights}


\end{frontmatter}



%
\section{Introduction} \label{sec1}
Humans are capable of perceiving the canonical colors of objects under various illumination conditions, a visual system feature known as color constancy.
In computer vision, the color constancy models aim to estimate illuminant color, followed by calibrating color-biased images to ensure accuracy in downstream tasks like object detection and segmentation.
In recent years, the rise of the Deep Neural Network~(DNN) has driven substantial advancements in color constancy research.
However, extensive research~\cite{goodfellow2014explaining, kurakin2018adversarial, madry2018towards, carlini2017towards} has revealed inherent vulnerabilities in DNN, where even minor perturbations can cause significant degradation in performance. Unfortunately, the robustness of DNN-driven Color Constancy (DNNCC) has largely been overlooked, resulting in insufficient exploration of potential perturbations that could limit the performance of DNNCC models.

\par DNNCC primarily focuses on the chromaticity of the illuminant rather than its brightness levels. Nonetheless, brightness remains a critical factor in color constancy tasks, providing valuable cues for identifying key features such as light sources, specular reflections, and achromatic surfaces, ultimately enhancing the accuracy of illuminant estimation. For instance, Land et al.~\cite{land1977retinex} proposed the white patch hypothesis, which posits that the maximum response in an image’s RGB channels is produced by surfaces that perfectly reflect incident light, thus, the pixels with the highest brightness can be considered indicative of the illuminant's color. 
Additionally, Bianco et al.~\cite{bianco2019quasi} utilized an image-to-image translation network to identify achromatic surfaces from brightness maps for the purpose of illuminant estimation. Moreover, numerous DNNCC models~\cite{hu2017fc4,yu2020cascading,lo2021clcc} perform end-to-end illuminant estimation using RGB images, implicitly leveraging brightness information as part of the process.
\par
\begin{figure*}[t]
  \includegraphics[width=1\linewidth]{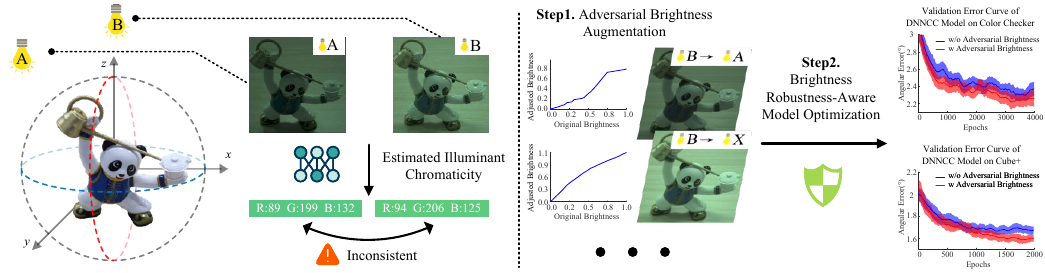}
   \vspace{-0.7cm}
  \caption{
The DNNCC model is sensitive to brightness variation. On the~\textbf{left}, two light sources~(A and B) with identical chromaticity but different positions create varying brightness, leading to inconsistent chromaticity estimates. This suggests that brightness variation could be a potential cause of degradation in the DNNCC model. On the~\textbf{right}, we propose a Brightness Robustness Enhancement~(BRE) strategy, which mitigates these effects by adversarially augmenting high-risk brightness images and incorporating robustness-aware optimization during training, thereby significantly reducing illuminant estimation errors across multiple datasets.}
  \label{fig1}
\end{figure*}

 Although numerous studies have explicitly or implicitly employed brightness information to estimate illuminant chromaticity, the potential influence of brightness variability on DNNCC remains largely unexamined. In the real-world, scenes inherently exhibit heterogeneous brightness distributions, shaped by a complex interplay of factors. These factors encompass illuminant characteristics (e.g., intensity, distance, and angle of incidence), intrinsic object properties (e.g., their three-dimensional structure and surface reflectance properties), as well as camera parameters (e.g., shutter speed and aperture size), among others. Brightness variation induced by these multifaceted factors may significantly impact the performance of DNNCC models.
The left part of Figure~\ref{fig1} illustrates a simulated experiment in which an identical scene is illuminated under two illumination conditions with equivalent chromaticity but different brightness levels. Notably, the predicted illuminant estimates by the DNNCC models were inconsistent.
Given that color constancy datasets are typically collected from real-world scenes with varied brightness, these inconsistencies could undermine the performance of DNNCC models.

Motivated by these observations and analyses, we propose the following two unresolved questions in color constancy tasks:
\begin{itemize}
    \item \textit{How substantially does brightness variation impact the performance of DNNCC models?}
    \item \textit{How to enhance DNNCC’s robustness against brightness variation?}
\end{itemize}
\par 
In this paper, we begin by addressing the first question through a systematic evaluation of DNNCC's brightness robustness. 
We develop a new color constancy dataset specifically designed to analyze the effects of brightness variation. Specifically, to mitigate the influence of non-brightness factors in imaging, we employed photorealistic rendering techniques to generate two sets of images with identical illuminant chromaticity but differing brightness levels. By utilizing these two sets independently for training and testing, we effectively isolate the confounding effects of non-brightness variables, allowing for a rigorous quantitative assessment of the influence of brightness alterations on the performance of the DNNCC model.
\par Subsequently, we propose a brightness robustness enhancement strategy for the DNNCC model, called BRE, to tackle the second issue outlined above. 
BRE consists of two components: adversarial brightness augmentation and brightness-robustness-aware model optimization. The primary component, adversarial brightness augmentation, aims to identify potentially risky brightness variation. By explicitly adjusting the parametric brightness curve to simulate changes in image brightness, the optimization problem of adversarial brightness augmentation can be defined as perturbing the brightness curve parameters to shift the DNNCC model’s output away from the true chromaticity value of the illuminant. In particular, the perturbation magnitude is adaptively calibrated in accordance with the model's brightness robustness, facilitating the generalization of adversarial brightness augmentation across various datasets and models.
Once the augmented images are acquired, BRE employs them in a brightness-robustness-aware optimization to improve the model's robustness against brightness variation, incorporating two main components: adversarial brightness training and brightness contrastive learning. Adversarial brightness training formulates the training process as a max-min optimization problem. In this process, adversarial brightness images are generated by maximizing the model's loss, then integrated into the model training to minimize the model's loss and improve its robustness against brightness variation. In addition, brightness contrastive learning is designed to align the feature representations of augmented and original images, thereby encouraging DNNCC models to extract brightness-invariant features.

\par 
It is worth noting that the proposed BRE aims to enhance the DNNCC model’s capacity to handle brightness variation in real-world scenarios, rather than defending against malicious attacks, as is typical in traditional AI security research.
To the best of our knowledge, this is the first work to leverage adversarial attacks and defenses to enhance DNNCC’s performance in color constancy tasks. As illustrated in the lower right corner of Figure~\ref{fig1}, the proposed BRE effectively improves the DNNCC model's performance on clean datasets, highlighting the significance of enhancing brightness robustness for the DNNCC model.

\par Our contributions can be summarized as follows.
\begin{itemize}
    \item 
    We first propose the concept of “brightness vulnerability” for color constancy. By systematically evaluating the DNNCC model's robustness to brightness variations, we identified it as a potential cause of model degradation.
    \item 
    We propose BRE, an incremental plug-in designed to enhance the brightness robustness of existing DNNCC models. BRE adaptively generates adversarially augmented brightness images via parameterized brightness curves, and further integrates brightness-robustness-aware optimization to enhance the performance of DNNCC models under diverse brightness variations.
    \item 
    Comprehensive evaluations on two real-world color constancy datasets indicate that the proposed BRE consistently enhances the performance of mainstream DNNCC models, without incurring additional computational 
    overhead during the testing phase.
\end{itemize}

\section{Related Work} \label{sec2}
\subsection{Image Formation for Color Constancy} \label{sec2.1}
\par Based on the Dichromatic Reflection Model~\cite{klinker1990physical}, the image $\boldsymbol{f}(\boldsymbol{x}) = \left( f_R(x), f_G(x), f_B(x) \right)^\top$ is determined by the scene illuminant $I(\boldsymbol{x}, \lambda)$, the camera sensor response function $\boldsymbol{\rho}(\lambda) = \left( \rho_R(\lambda), \rho_G(\lambda), \rho_B(\lambda) \right)^\top$, the surface reflectance $S(\boldsymbol{x}, \lambda)$, and the parameters $m_b(\boldsymbol{x})$ and $m_s(\boldsymbol{x})$ which represent body reflection and surface reflection, respectively~\cite{gijsenij2011computational}, such that:
\begin{equation} \label{eq1}
\begin{aligned}
  f_c(\boldsymbol{x}) &= m_b(\boldsymbol{x}) \int_\omega I(\boldsymbol{x}, \lambda) \rho_c(\lambda) S(\boldsymbol{x}, \lambda) \, d\lambda \\
  &\quad\quad + m_s(\boldsymbol{x}) \int_\omega I(\boldsymbol{x}, \lambda) \rho_c(\lambda) \, d\lambda,
\end{aligned}
\end{equation}
where $c \in \{R, G, B\}$, $\lambda$ denotes the illuminant wavelength, and $\omega$ represents the visible wavelength range. The scaling factors $m_b(\boldsymbol{x})$ and $m_s(\boldsymbol{x})$ depend on the viewing angle, illuminant direction, and surface orientation~\cite{van2006coloring}. 
The objective of the color constancy model is to estimate the illuminant chromaticity from the color-biased image $\boldsymbol{f}(\boldsymbol{x})$, denoted as $\boldsymbol{e}(\boldsymbol{x})$:
\begin{equation} \label{eq2}
\boldsymbol{e}(\boldsymbol{x}) = \left(e_R(\boldsymbol{x}), e_G(\boldsymbol{x}), e_B(\boldsymbol{x})\right)^\top = \int_\omega I(\boldsymbol{x}, \lambda) \rho_c(\lambda) \, d\lambda.
\end{equation}

\subsection{Color Constancy} \label{sec2.2}
Color constancy methods typically assume that the scene is illuminated by a single illuminant. Under this assumption, the objective of color constancy model is to predict the $R$, $G$, and $B$ values for each image. In this paper, we follow the assumption and categorize these methods into two groups—DNN-based and non-DNN-based—providing an overview of each.

\subsubsection{Non-DNN Methods} \label{sec2.2.1}
Early non-DNN color constancy methods typically estimate the illuminant through statistical assumptions, including the gray-world~\cite{buchsbaum1980spatial}, white-patch~\cite{land1977retinex}, and gray-edge assumptions~\cite{van2007edge}, among others. These statistical assumptions can be encompassed within a unified framework~\cite{van2007edge}:
\begin{equation} \label{eq3}
\left( \int \left| \frac{\partial^n \boldsymbol{f}^{\sigma}(\boldsymbol{x})}{\partial \boldsymbol{x}^n} \right|^p \, d\boldsymbol{x} \right)^{\frac{1}{p}} = k \cdot \boldsymbol{e}^{n,p,\sigma},
\end{equation}
where the variable \( n \) denotes the order of the derivative, \( \sigma \) represents the standard deviation of a Gaussian filter, with \( \boldsymbol{f}^{\sigma}(\boldsymbol{x}) = G^{\sigma} * \boldsymbol{f}(\boldsymbol{x}) \), and \( p \) specifies the order of the Minkowski norm. 
Eq.~\ref{eq3} assumes a fixed ratio \( k \) between image statistics and the illuminant, allowing the illuminant value to be estimated directly from image statistics.
Similarly, some learning-based non-DNN methods~\cite{li2023ranking, xie2022improving, xie2023camera} also build machine learning models based on images statistical assumptions. Moreover, additional methods seek to develop models through the lenses of gamut mapping~\cite{forsyth1990novel, gijsenij2010generalized}, Bayesian inference~\cite{gehler2008bayesian}, regression trees~\cite{cheng2015effective}, frequency domain~\cite{barron2017fast}, and so on.

\subsubsection{DNN-based Methods} \label{sec2.2.2}
\par With the advancement of deep learning, Deep Neural Network-driven Color Constancy~(DNNCC) has attracted significant attention from researchers.
Bianco et al.~\cite{bianco2015color} first introduced CNN for illuminant chromaticity estimation. Given the limited number of training samples typically found in color constancy datasets, they divided images into patches to alleviate data insufficiency.
Additionally, Yu et al.~\cite{yu2020cascading} proposed a cascading architecture to capture dependencies between light source hypotheses, facilitating a coarse-to-fine estimation. Furthermore, both IGTN~\cite{xu2020end} and CLCC~\cite{lo2021clcc} emphasized that the performance of DNNCC is highly sensitive to variations in scene content. They utilize metric learning and contrastive learning, respectively, to focus on illuminant-dependent features. 
\par Given the ill-posed nature of the color constancy problem, researchers have explored various strategies to handle the inherent ambiguities. For instance, Shi et al.~\cite{shi2016deep} developed DS-Net to mitigate ambiguities from unknown reflections and object appearances. Similarly, Song et al.~\cite{hu2017fc4} framed color constancy as a grouped regression problem, generating multiple possible illuminant solutions to tackle these ambiguities. Moreover, Hu et al.~\cite{hu2017fc4} introduced a confidence-weighted layer to identify key regions crucial for illuminant estimation, thereby reducing the influence of uncertainty. Recently, Buzzelli et al.~\cite{buzzelli2024uncertainty} proposed three uncertainty estimation strategies for color constancy and developed cascaded methods based on the estimated uncertainties.
\par Unlike the aforementioned studies that directly estimate illuminant chromaticity, Hernandez-Juarez et al.~\cite{hernandez2020multi} define color constancy as a classification task instead of a regression task, determining whether candidate illuminant correctly adjusts the image. 
Additionally, Bianco et al.~\cite{bianco2019quasi} employed a brightness map to identify achromatic regions, directly inferring the illuminant chromaticity.
\par Alongside the utilization of RGB images as input, recent studies have investigated the integration of additional data sources, such as the degree of linear polarization~\cite{xing2022point}, point clouds~\cite{xing2022point}, and multiple rear-facing cameras~\cite{abdelhamed2021leveraging}, to improve illuminant estimation accuracy. These methods have shown significant improvements in accuracy, attributed to the integration of additional data sources. However, reliance on specialized hardware limits their accessibility in broader applications.
\par Beyond improving the accuracy of DNNCC models, some studies have focused on reducing their computational complexity. Buzzelli et al.~\cite{buzzelli2024convolutional} proposed a convolutional model based on low-level image features to enable efficient illuminant estimation through image statistics. Similarly, Domislović et al.~\cite{domislovic2022one} introduced One-Net, a lightweight network with five 1$\times$1 kernel convolutional layers, designed to leverage low-level image statistics. Moreover, Laakom et al.~\cite{laakom2020bag} introduced BoCF, which achieves a reduction in the number of model parameters required for illuminant estimation through bag-of-features pooling.
\par Aside from illuminant estimation, another key challenge in color constancy is generalizing across varied camera sensors, as spectral sensitivity differences cause shifts in illuminant distributions. Xiao et al.~\cite{xiao2020multi} and Zhang et al.~\cite{zhang2022domain} addressed this with multi-domain and domain-adversarial learning to derive shared features from different sensors. Alternatively, Afifi et al.~\cite{afifi2021cross} advanced this by leveraging unlabeled test-phase images for dynamic inference of sensor spectral sensitivity profiles. Tang et al.~\cite{tang2022transfer} introduced a statistical approach to convert sensor-specific illuminant labels to a sensor-agnostic format, promoting model generalization across diverse sensors.

\subsection{Data Augmentation for DNNCC} \label{sec2.3}
\par Data augmentation diversifies the training data through various transformations. Common geometric transformations, such as random flipping, cropping, and rotating, are also used for color constancy~\cite{hu2017fc4, yu2020cascading, xu2020end, lo2021clcc, hernandez2020multi}. Nevertheless, due to the high sensitivity to color variations, most color transformations, such as channel dropping and swapping, are unsuitable for color constancy. The most prevalent color augmentation technique for color constancy involves linearly scaling the illuminant and adjusting image colors simultaneously~\cite{hu2017fc4}, thus creating richer scene-light combinations. A more precise method~\cite{lo2021clcc} identifies 24-color swatch values and uses a transformation matrix to swap illuminants between images.
Although the aforementioned data augmentation significantly enhanced the performance of DNNCC, these strategies did not explicitly consider brightness transformations. Our study reveals a general issue with the brightness robustness of DNNCC models. To tackle this challenge, we propose an adversarial brightness augmentation strategy. This approach effectively mitigates performance degradation caused by brightness sensitivity by learning from high-risk brightness transformations, demonstrating its effectiveness as an innovative augmentation method for DNNCC models.
\subsection{Adversarial Attack and Defense} \label{sec2.4}
DNN have achieved state-of-the-art performance across a wide range of tasks. However, extensive research has revealed that DNN are highly susceptible to adversarial examples. Szegedy et al.~\cite{szegedy2013intriguing} were the first to introduce the concept of adversarial examples, employing the L-BFGS method to generate small perturbations that, when added to clean samples, cause DNN to produce incorrect predictions with high confidence. Subsequently, Goodfellow et al.~\cite{goodfellow2014explaining} proposed the Fast Gradient Sign Method (FGSM) as a more efficient approach to generating adversarial examples:
\begin{equation}~\label{eq4}
\begin{gathered}
    \delta = \varepsilon sign\left( {{\nabla }_{\boldsymbol{x}}}\mathcal{J} \left( \mathcal{M}\left( \boldsymbol{f}(\boldsymbol{x}) \right),y \right) \right),
\end{gathered}
\end{equation}
where \(\delta\) is an additive perturbation, defined as the product of the scalar \(\varepsilon\) and the sign of the gradient \(\nabla_{\boldsymbol{x}}\) of the loss function \(\mathcal{J}(\mathcal{M}(\boldsymbol{f}(\boldsymbol{x})), y)\), where $\mathcal{J}$, $\mathcal{M}$, $y$, and $\boldsymbol{x}$ represent the loss function, DNN model, ground truth, and clean sample, respectively, with $\boldsymbol{f}(\boldsymbol{x}) = \left( f_R{(\boldsymbol{x})}, f_G{(\boldsymbol{x})}, f_B{(\boldsymbol{x})} \right)^\top$. The scalar \(\varepsilon\) ensures the perturbation to be minimal but effective to mislead the model. Similar to FGSM, numerous attack variants such as PGD~\cite{madry2018towards} and C\&W~\cite{carlini2017towards} have been proposed. Additionally, some research has shifted focus from pursuing powerful attacks to expanding perturbation types through techniques such as altering color temperature~\cite{afifi2019else} and retouching~\cite{xie2024retouchuaa}
\par The success of adversarial attacks has driven advancements in defense research. To improve the robustness of DNN against adversarial examples, several methods have been proposed, including adversarial training~\cite{madry2018towards}, defensive distillation~\cite{papernot2016distillation}, and gradient regularization~\cite{yan2018deep}, etc. Additionally, some defences methods focus on preprocessing to eliminate adversarial perturbations, such as JPEG compression~\cite{dziugaite2016study}, as well as denoising using diffusion models~\cite{nie2022diffusion}.
Additionally, some research has focused on detecting adversarial samples during the testing phase. For instance, Metzen et al.\cite{metzen2017detecting} trained auxiliary models to identify adversarial inputs by using intermediate layer outputs as features. Gao et al.\cite{gao2023detecting} developed an autoencoder-based model robust to adversarial perturbations, detecting adversarial samples by comparing prediction consistency with the protected model.
Gao et al.~\cite{gao2023detecting} developed an autoencoder-based model that is robust to adversarial perturbations, detecting adversarial samples by comparing prediction consistency with the protected model.


\section{Brightness Robustness Evaluation for DNNCC} \label{sec3}
\par We start with conducting quantitative evaluations of the impact of brightness variation on DNNCC models, providing valuable insights into their sensitivity to such changes. This is accomplished by separately assessing the model's illuminant estimation performance on the training and testing sets, which differ only in terms of brightness variation. A DNNCC model with strong brightness robustness is expected to maintain consistent performance levels between training and testing phases.
\par
\begin{figure}[t]
  \includegraphics[width=1\linewidth]{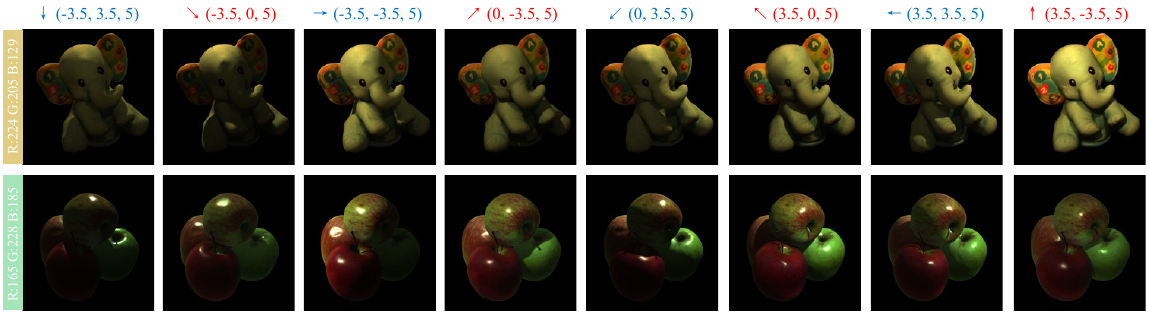}
   \vspace{-0.7cm}
  \caption{Visualization of the synthetic dataset. The colored bar on the left shows illuminant labels, with arrows and coordinates above the image indicating the illumination direction and world coordinate position.  \textcolor{red}{Red} and  \textcolor{blue}{blue} represent the training and test sets, respectively.}
  \label{fig2}
\end{figure}

 The primary challenge in evaluating the brightness robustness of DNNCC models is ensuring consistency in non-brightness factors—such as scene content, illuminant chromaticity, and camera viewpoint—across both the training and test datasets. However, no existing dataset fully meets these specific requirements. To address this, we have developed a synthetic dataset using Blender Cycles rendering engine. This dataset comprises 800 images generated from 20 distinct scenes, each captured from a fixed camera viewpoint.
For each scene, we selected five illuminant labels. Each illuminant label was used to define the chromaticity of both the point light source and the ambient light source. These two light sources were combined to render the images. To vary the brightness of the images, we randomly adjusted the brightness of the ambient light source, while changing the position of the point light source and directing its light toward the center of the scene. Under the same illuminant label setting, each point light source position was rendered once. This process resulted in a total of 20 scenes $\times$ 5 illuminant chromaticities $\times$ 8 point light positions = 800 images.
Figure~\ref{fig2} illustrates the partially visualized results. We split the training and test sets into 1:1 partitions according to eight evenly spaced point light source positions.
\par To comprehensively assess the brightness robustness of DNNCC models, we selected six recently published or high-performance DNNCC models: FC4~\cite{hu2017fc4}, Quasi-UCC~\cite{bianco2019quasi}, C4~\cite{yu2020cascading}, EIL-Net~\cite{cun2022learning}, ECF~\cite{buzzelli2024convolutional} and GC\textsuperscript{3}~\cite{zhengguang2025gc3}.
For all evaluated models, we set the batch size to 16 and the training epochs to 4000, with other settings maintained according to the original papers. To mitigate overfitting due to the small size of the proposed brightness robustness dataset, we trained for 2000 epochs on the Cube+ dataset~\cite{banic2017unsupervised} with standard data augmentation techniques (including cropping, rotation, and variations in light sources) and selected the best model on the ColorChecker dataset~\cite{gehler2008bayesian} as the pre-trained model. For the proposed brightness robustness dataset, all data augmentation was disabled to prevent altering non-brightness factors in both training and test sets. Table~\ref{tab1} shows the illuminant estimation errors for the DNNCC models on both the training and test sets.

\begin{table}[t]
\centering
\fontsize{14}{15}\selectfont 
\renewcommand{\arraystretch}{1.4}
\begin{adjustbox}{max width=\textwidth}
\begin{tabular}{l|>{\centering\arraybackslash}p{2.7cm}|>{\centering\arraybackslash}p{2.7cm}|>{\centering\arraybackslash}p{2.7cm}|>{\centering\arraybackslash}p{2.7cm}|>{\centering\arraybackslash}p{2.7cm}|>{\centering\arraybackslash}p{2.7cm}} 
\specialrule{1pt}{0pt}{0pt} \noalign{\smallskip}
\multirow{1}{*}{\textbf{DNNCC Models}} & \textbf{FC4} & \textbf{Quasi-UCC} & \textbf{C4}  & \textbf{EIL-Net} & \textbf{ECF} & \textbf{GC\textsuperscript{3}} \\ \noalign{\smallskip} \hline \noalign{\smallskip}
\textbf{Training set}                  &     0.13        &      0.79      & 0.17               &   0.12            &   0.56                & 0.33                \\
\textbf{Test set}                      &      0.81       &       2.96     & 1.16              &   0.99            & 2.74                   & 2.00                   \\ \noalign{\smallskip} \hline \noalign{\smallskip}
\textbf{Error Increase}            &  521.15\%       &  274.68\%       &  582.35\%         &  725.00\%          &  389.29\%              &  506.06\%              \\ \noalign{\smallskip} \specialrule{1pt}{0pt}{0pt}
\end{tabular}
\end{adjustbox}
 \vspace{-0.3cm}
\caption{Evaluation of Brightness Robustness: The performance of various DNNCC models in illuminant estimation on the training and test sets of the brightness robustness dataset is evaluated using the mean angular error. A lower angular error indicates better performance. To reduce the impact of outliers, the angular error is smoothed using a moving average over epochs with a window size of 500. The percentage increase in error is calculated by comparing the angular error between the test set and the training set.}
\label{tab1}
 \vspace{0.7cm}
\end{table}

\begin{figure}[t]
  \includegraphics[width=1\linewidth]{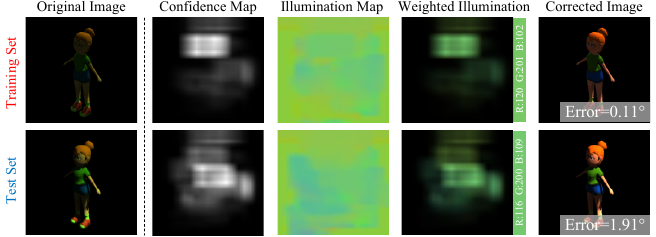}
 \vspace{-0.7cm}
  \centering
  \caption{Visualization of brightness variation impact on the FC4 model. The estimated illuminant and error are shown on the right of the fourth and fifth column images, respectively.}
  \label{fig3}
\end{figure}
    
\par As shown in Table~\ref{tab1}, the illuminant estimation errors for all DNNCC models on the test set are markedly higher than those on the training set, despite the presence of only brightness variation between the two datasets. This reveals a limitation in the robustness of DNNCC models when exposed to brightness changes. Furthermore, Figure~\ref{fig3} illustrates the impact of brightness variation on the FC4 model. Significant discrepancies are observed in the illumination map predicted by the FC4 model across different brightness levels. It is notable that the FC4 model also generates confidence maps to support weighted illuminant estimation. However, these confidence maps display considerable changes as well, suggesting that brightness variation may degrade the final illuminant estimation performance by causing DNNCC models to incorrectly focus on unstable regions.
brightness robustness enhancement
\section{Brightness Robustness Enhancement in DNNCC Models} \label{sec4}
\par
\begin{figure}[t]
  \includegraphics[width=1\linewidth]{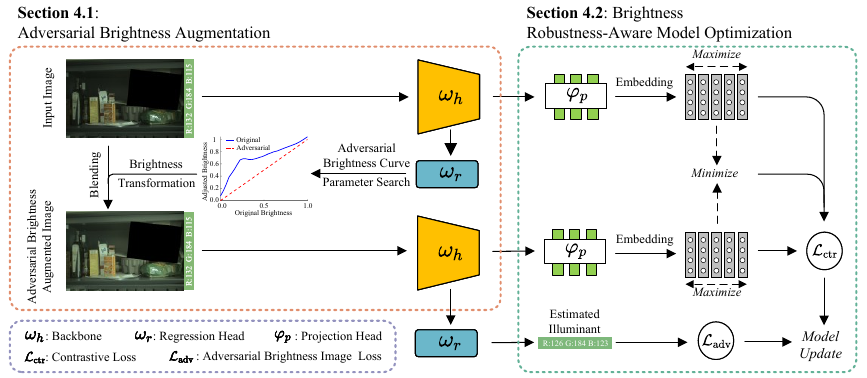}
  \vspace{-0.7cm}
  \caption{An overview of the proposed brightness robustness improvement framework}
  \label{fig4}
\end{figure}

 In light of the observed limitations in brightness variation within the DNNCC model, this section aims to enhance the model's robustness to brightness, thereby improving its illuminant estimation accuracy. The proposed strategy, \textbf{BRE}, consists of two main components: (1) \textbf{adversarial brightness augmentation}, and (2) \textbf{brightness-robustness-aware model optimization}.
It is noteworthy that the proposed framework can be seamlessly integrated into existing DNNCC models as an additional module without modifying the architecture, while also introducing no extra computational overhead during the testing phase. Figure~\ref{fig4} provides an overview of the proposed framework. 

\subsection{Adversarial Brightness Augmentation} \label{sec4.1}
\subsubsection{Brightness Transformation Modeling with Parameterized Brightness Curves} \label{sec4.1.1}
\par In this section, we focus on modeling brightness transformations to enable controlled brightness adjustments, which serve as the basis for adversarial brightness augmentation. For the transformation to be practical, it must satisfy two key requirements: differentiability, allowing gradient descent optimization of adversarial brightness parameters during the adversarial brightness augmentation phase; and efficiency, enabling the generation of diverse brightness images without substantially increasing training time.
\par Motivated by these considerations, we adopt a piecewise linear function to parameterize brightness curve. This renders the parametric brightness curve differentiable and computationally efficient, thereby facilitating its integration into our framework for adversarial brightness parameter search and real-time brightness manipulation. The mathematical model for brightness filtering, denoted as $\mathcal{F}$, based on this parameterized brightness curve, is given as follows:
\begin{equation} \label{eq5}
\begin{aligned}
  {\boldsymbol{f}(\boldsymbol{x})'} &= \mathcal{F}({{{u}(\boldsymbol{x})}}, \boldsymbol{\theta}) \\
  &= {\frac{\boldsymbol{f}(\boldsymbol{x})}{{u}(\boldsymbol{x})}} \cdot EnBright({u}(\boldsymbol{x}), \boldsymbol{\theta})  \\
  &= {\frac{\boldsymbol{f}(\boldsymbol{x})}{{u}(\boldsymbol{x})}} \cdot \frac{1}{T_{L}} \sum_{j=0}^{L-1} \max(L \cdot {{u}(\boldsymbol{x})} - j, 0) \cdot \theta_{j},
\end{aligned}
\end{equation}
where $\boldsymbol{f}(\boldsymbol{x})'$ denotes the brightness-adjusted image. $u(\boldsymbol{x}) = \mathcal{N}(\frac{1}{3} 
(f_R(\boldsymbol{x})$ $+ f_G(\boldsymbol{x}) + f_B(\boldsymbol{x})))$ represents the brightness map and $\mathcal{N}$ denotes the normalization function. $\boldsymbol{\theta} = (\theta_{1}, \theta_{2}, \ldots, \theta_{L-1})$ represents the parameter set of the brightness curve, with each $\theta$ initialized to $1/L$. The parameterize brightness curve is controlled by a set of points $\left({j}/{L}, {T_{j}}/{T_{L}}\right)$, where $T_{j} = \sum_{i=0}^{j-1} \theta_{i}$ represents the cumulative histogram counts up to the $j$-th level. Considering the accuracy of curve simulation and computational cost, we set $L=32$ to approximate $EnBright$ as a curve in our experiment.

\subsubsection{Adversarial Brightness Parameter Optimization} \label{sec4.1.2}
\par The core motivation of the proposed BRE is to enhance the brightness robustness of the DNNCC model by augmenting images with diverse brightness levels. While simple global random intensity adjustments provide a broad range of brightness variations, it lacks model-specific optimization and fails to pinpoint high-risk brightness levels, thereby limiting the improvements in brightness robustness.
In contrast, we introduce adversarial brightness parameter optimization for image augmentation, which identifies challenging brightness variation by maximizing the model’s loss function. These variations reveal the model's instability under varying lighting conditions, effectively guiding it to learn brightness-robust features. 
Building upon the parametrized brightness curve \( \mathcal{F} \) defined in Section~\ref{sec4.1.1}, the adversarial brightness parameters
\( \boldsymbol{\theta}^* \) that define the optimization objective for brightness transformations can be mathematically modeled as follows:
\begin{equation} \label{eq6}
\begin{aligned}
    \boldsymbol{\theta}^* = \arg \max_{\boldsymbol{\theta}} \mathcal{J}(\mathcal{M}(\mathcal{F}(\boldsymbol{f}(\boldsymbol{x}), \boldsymbol{\theta}); \boldsymbol{w}), \boldsymbol{\ell}),
\end{aligned}
\end{equation}
where $\mathcal{M}(\mathcal{F}(\boldsymbol{f}(\boldsymbol{x}), \boldsymbol{\theta}); \boldsymbol{w})$ denotes the normalized illuminant estimated by the DNNCC model $\mathcal{M}$ for the image adjusted by the parametrized brightness curve $\mathcal{F}$, with $\boldsymbol{x}$ as the clean image, $\boldsymbol{\theta}$ as the brightness curve parameter, and $\boldsymbol{w}$ as the model parameters. The symbol $\boldsymbol{\ell}$ denotes the ground truth normalized illuminant. The loss function $\mathcal{J}$, commonly used in color constancy, is defined as the angular error $\mathcal{J}=\frac{180}{\pi }\arccos \left( \mathcal{M}(\mathcal{F}(\boldsymbol{f}(\boldsymbol{x}), \boldsymbol{\theta}); \boldsymbol{w})\cdot \boldsymbol{\ell} \right)$. The Eq.~\ref{eq6} aims to identify adversarial brightness parameters $\boldsymbol{\theta}^*$ that maximize the prediction error of the DNNCC model on brightness-adjusted images. 
\par In white-box scenario, Eq.~\ref{eq6} can be efficiently optimized by the basic iterative method~\cite{kurakin2018adversarial}:
\begin{equation} \label{eq7}
\begin{aligned}
\boldsymbol{\theta}^* = \boldsymbol{\theta} - \alpha \frac{\boldsymbol{g}}{\|\boldsymbol{g}\|_2}, \quad
\text{where } \boldsymbol{g} = \nabla_{\boldsymbol{\theta}} \mathcal{J}(\mathcal{M}(\mathcal{F}(\boldsymbol{f}(\boldsymbol{x}), \boldsymbol{\theta}); \boldsymbol{w}), \boldsymbol{\ell}).
\end{aligned}
\end{equation}
$\nabla_{\boldsymbol{\theta}}$ denotes the loss gradient with respect to the parameters $\boldsymbol{\theta}$, and $\alpha$ represents the step size controlling update magnitude during optimization.
\par The step size $\alpha$ in Eq.~\ref{eq7} is fixed as a constant that requires manual tuning, limiting its practicality in real-world applications. We address this challenge through a novel adaptive step size strategy that dynamically responds to brightness sensitivity. Intuitively, a DNNCC model exhibiting high brightness sensitivity ought to employ a larger step size $\alpha$ to explore a wider spectrum of brightness variations, and vice versa.
\par In the context of color constancy, the brightness sensitivity of a DNNCC model is primarily shaped by two critical factors. First, the model’s architecture imposes inherent inductive biases and feature extraction mechanisms, which in turn affect its ability to handle brightness changes. Second, training datasets encompassing diverse brightness distributions may introduce statistical coupling between brightness and chromaticity—the target of color constancy—leading the model to rely excessively on brightness cues and thereby increasing its sensitivity.
To quantify the brightness sensitivity of the DNNCC model, we introduce a statistical metric derived from the gradient of brightness curve parameters. In detail, before utilizing the gradient \(\boldsymbol{g}\), we perform outlier removal by replacing any values exceeding 1 or falling below -1 with the mean value, resulting in the processed gradient \(\boldsymbol{g}'\). For each training batch, we compute the $L_2$ norm of the gradient ${\|\boldsymbol{g}'\|_2}$ of model loss with respect to the brightness curve parameters $\boldsymbol{\theta}$ to measure the brightness sensitivity of the DNNCC model. Building on this, we leverage ${\|\boldsymbol{g}'\|_2}$ to adaptively determine the step size $\alpha$, which is further regulated by a momentum-based update mechanism across training batches. 
\begin{equation} \label{eq8}
\begin{aligned}
\alpha_t = m\alpha_{t-1} + (1-m)\frac{\|\boldsymbol{g}'\|_2}{10},
\end{aligned}
\end{equation}
where \( m \), the momentum coefficient set to 0.9 in our experiments, ensures stability of the step size updates during optimization, and 
$t$ represents the current training iteration. Following the computation of adversarial brightness parameters $\boldsymbol{\theta}^*$, the corresponding brightness transformation is applied via Eq.~\(\ref{eq5}\), generating adversarial brightness images $\mathcal{F}(\boldsymbol{f}(\boldsymbol{x}), \boldsymbol{\theta^*})$. Furthermore, to mitigate the overfitting of the DNNCC model to adversarial features~\cite{lee2020adversarial}, we perform blending between the clean image $\boldsymbol{f}(\boldsymbol{x})$, and the adversarial brightness image \( \boldsymbol{f}(\boldsymbol{x}^{'}) \) to obtain the augmented image $\hat{\boldsymbol{f}(\boldsymbol{x}^*)}$ for training:
\begin{equation} \label{eq9}
\begin{aligned} 
\hat{\boldsymbol{f}(\boldsymbol{x}^*)} = \lambda \boldsymbol{f}(\boldsymbol{x}^*) + (1 - \lambda) \boldsymbol{f}(\boldsymbol{x}), \quad \lambda \sim \mathcal{U}(0, 1),
\end{aligned}
\end{equation}
where $\lambda$ is a mixing coefficient sampled from a uniform distribution $\mathcal{U}(0, 1)$. The adversarial brightness augmentation algorithm is described in Algorithm~\ref{alg:adversarial_brightness_augmentation}.
\begin{algorithm}[t]
\footnotesize 
\caption{Adversarial Brightness Augmentation}
\label{alg:adversarial_brightness_augmentation}
\textbf{Input}: $\boldsymbol{f}(\boldsymbol{x})$: clean image, $\boldsymbol{\ell}$: ground truth, $\mathcal{F}$: brightness filter, $\mathcal{M}$: DNNCC model, $\boldsymbol{w}$: DNNCC model weights, $\mathcal{J}$: angle loss function \vspace{0.12cm} \\
\textbf{Parameters}: $\alpha_0=0.1$: initial step size, $m=0.9$: momentum coefficient, $n$: number of iterations, $L=32$: number of brightness curve parameters \vspace{0.12cm} \\
\textbf{Output}: $\hat{\boldsymbol{f}(\boldsymbol{x}^)}$: augmented brightness image \\
\begin{algorithmic}[1]
\FOR{each training step $t = 1$ to $n$} 
    \STATE Sample a mini-batch of training data $(\boldsymbol{f}_b(\boldsymbol{x}), \boldsymbol{\ell}_b)$ \vspace{0.12cm}
    \STATE Initialize brightness curve parameters $\boldsymbol{\theta} \leftarrow \frac{1}{L}$ \vspace{0.12cm}
    \STATE Apply brightness transformation $\boldsymbol{f}_b(\boldsymbol{x}') = \mathcal{F}(\boldsymbol{f}_b(\boldsymbol{x}), \boldsymbol{\theta})$ \vspace{0.12cm}
    \STATE Model predictions $\boldsymbol{\hat{\ell}_b} = \mathcal{M}(\boldsymbol{f}_b(\boldsymbol{x}'); \boldsymbol{w})$ \vspace{0.12cm}
    \STATE Calculate loss $\mathcal{J} = \frac{180}{\pi} \arccos\left(\boldsymbol{\hat{\ell}_b} \cdot \boldsymbol{\ell}_b\right)$ \vspace{0.12cm}
    \STATE Compute gradient $\boldsymbol{g} = \nabla_{\boldsymbol{\theta}} \mathcal{J}$, replace $|g_i| > 1$ in $\boldsymbol{g}$ with the mean value to get $\boldsymbol{g}'$ \vspace{0.12cm}
    \STATE Update step size $\alpha = m \cdot \alpha_{t-1} + (1 - m) \cdot \frac{\|\boldsymbol{g}'\|_2}{10}$ \vspace{0.12cm}
    \STATE Compute adversarial brightness parameters $\boldsymbol{\theta^*} = \boldsymbol{\theta} + \alpha_t \cdot \frac{\boldsymbol{g}'}{\|\boldsymbol{g}'\|_2}$ \vspace{0.12cm}
    \STATE Generate adversarial brightness image $\boldsymbol{f}_b(\boldsymbol{x^*}) = \mathcal{F}(\boldsymbol{f}_b(\boldsymbol{x}), \boldsymbol{\theta^*})$ \vspace{0.12cm}
    \STATE Sample interpolation coefficient $\lambda \sim \mathcal{U}(0, 1)$ \vspace{0.12cm}
    \STATE Generate augmented image $\hat{\boldsymbol{f}_b(\boldsymbol{x^*})} = \lambda \boldsymbol{f}_b(\boldsymbol{x^*}) + (1 - \lambda) \boldsymbol{f}_b(\boldsymbol{x})$ \vspace{0.12cm}
\ENDFOR
\STATE \textbf{Return} $\hat{\boldsymbol{f}_b(\boldsymbol{x}^*)}$
\end{algorithmic}
\end{algorithm}


\subsection{brightness-robustness-aware Model Optimization} \label{sec4.2}
In this section, we focus on enhancing the DNNCC model's robustness to brightness variation through joint adversarial and contrastive loss optimization. Building upon the adversarial brightness augmentation strategy introduced in Section \ref{sec4.2}, the core training process of the DNNCC model can be modeled as a min-max problem~\cite{madry2018towards}:
\begin{equation} \label{eq10}
\begin{aligned}
\min_{\boldsymbol{w}} \mathbb{E}_{(\boldsymbol{x}, \boldsymbol{\ell}) \sim \mathcal{D}} 
\left[ 
    \max_{\boldsymbol{\theta}} \mathcal{J}\left(\mathcal{M}\left(\mathcal{F}(\boldsymbol{f}(\boldsymbol{x}), \boldsymbol{\theta}); \boldsymbol{w}\right), \boldsymbol{\ell}\right) 
\right],
\end{aligned}
\end{equation}
where $\mathcal{D}$ represents the training data distribution, $\boldsymbol{w}$ denotes the DNNCC model weights, and the remaining symbols are consistent with Eq.~\ref{eq6}. The inner maximization problem in Eq.~\ref{eq10} identifies adversarial brightness images that maximize the DNNCC model's loss, while the outer minimization problem adjusts the model parameters $\boldsymbol{w}$ to minimize the loss on these images.
\par Furthermore, we construct contrastive pairs to explicitly learn feature representations that are both discriminative and invariant to brightness within the embedding space. In contrast to conventional self-supervised contrastive learning approaches, our method utilizes supervised label information to adversarially generate challenging brightness-augmented images, combined with clean samples to form contrastive pairs.
Specifically, given the original image set $\boldsymbol{f}(\boldsymbol{x})$ and the corresponding set of images $\boldsymbol{f}^*(\boldsymbol{x})$ with adversarial brightness augmentation, we extract the high-dimensional feature maps $\boldsymbol{h}$ and $\boldsymbol{h}^*$ from the backbone network $h_\phi(\cdot)$ of the DNNCC model for both sets. The extracted feature maps are then projected into embeddings $\boldsymbol{z}$ and $\boldsymbol{z}^*$ using an MLP projection head $g_\psi(\cdot)$. Let $z_i \in \boldsymbol{z}$ and $z_i^* \in \boldsymbol{z}^*$, where $i \neq j$. In the contrastive loss, positive sample pairs are defined as embedding pairs $(z_i, z_i^*)$ from the original image and its brightness-augmented counterpart. Negative sample pairs are derived from two sources: embedding pairs from the original image~$(z_i, z_j)$, and embedding pairs between the original and augmented image $(z_i, z_j^*)$. 
We employ the InfoNCE loss to jointly optimize the similarity between positive and negative embedding pairs, as defined by the following:
\begin{equation} \label{eq11}
\begin{aligned}
\mathcal{L}_{\text{con}} = -\frac{1}{N} \sum_{i=1}^{N} \log \frac{\exp(\text{sim}({z}_i, {z}_i^*) / \tau)}{\exp(\text{sim}({z}_i, {z}_i^*) / \tau) + \sum_{{z} \in \mathcal{N}_i} \exp(\text{sim}({z}_i, {z}_j) / \tau)},
\end{aligned}
\end{equation}
where $\mathcal{N}_i = \left\{ \boldsymbol{z}_j \mid j \neq i \text{ and } \boldsymbol{z}_j \in \boldsymbol{z} \right\} \cup \left\{ \boldsymbol{z}_j^* \mid j \neq i \text{ and } \boldsymbol{z}_j^* \in \boldsymbol{z}^* \right\}$, $N$ denotes the batch size, and $\tau$ represents the temperature parameter, which controls the smoothness of the similarity between sample pairs. We set $\tau=1$ to ensure that the learned features can smoothly adapt to the continuous label variations inherent in this regression-based color constancy task. $\text{sim}(\cdot)$ denotes the cosine similarity between the embedding pairs. Eq.~\ref{eq11} encourages the DNNCC model to extract brightness-robust discriminative features by increasing the similarity within positive pairs while reducing that within negative pairs.
\par Finally, by integrating Eq.~\ref{eq10} and Eq.~\ref{eq11}, we propose the joint loss as:
\begin{equation} \label{eq12}
\mathcal{L}_{\text{joint}} = \lambda_{adv} \mathcal{L}_{\text{adv}} + \lambda_{ctr} \mathcal{L}_{\text{ctr}},
\end{equation}
where the adversarial loss $\mathcal{L}_{\text{adv}} = \mathcal{J}\left(\mathcal{M}\left(\mathcal{F}(\boldsymbol{f}(\boldsymbol{x}), \boldsymbol{\theta}^*); \boldsymbol{w}\right), \boldsymbol{\ell}\right)$, calculated from augmented images generated with adversarial brightness parameters $\boldsymbol{\theta}^*$, is minimized to enhance the DNNCC model's robustness against brightness variation. Moreover, the contrastive loss $\mathcal{L}_{\text{ctr}}$ further enhances the model's brightness robustness by minimizing the distance between the feature representations of the original and brightness-augmented images in the embedding space.


\section{Experiment} \label{sec5}
In this section, we evaluate the effectiveness of the proposed strategy in improving performance on color constancy datasets by enhancing brightness robustness. The experimental setup is as follows:
\vspace{0.25cm}
\par \noindent \textbf{Datasets and Metric.}
Two publicly available color constancy datasets— ColorChecker \cite{gehler2008bayesian} and Cube+\cite{banic2017unsupervised}—along with the brightness robustness color constancy dataset introduced in Section\ref{sec3}, were utilized. Each image in these datasets is accompanied by ground-truth illuminant chromaticity. The illuminant estimation error for each image of the DNNCC model was evaluated using the angular error function $\mathcal{J}$. 
Following prior work~\cite{hu2017fc4}, we conducted three-fold cross-validation on both the ColorChecker and Cube+ datasets. For the brightness robustness color constancy dataset, we employed the cross-dataset evaluation protocol detailed in Section~\ref{sec3}, where models were trained on one dataset and evaluated on another, to quantitatively assess how the proposed BRE improves the DNNCC model's robustness to brightness variations.
To evaluate the performance of DNNCC models, we smoothed the test error metrics curve over epochs by applying a moving average with a window size of 500, thereby reducing the influence of outliers. From the resulting smoothed curves, we then calculated five metrics for the angular errors across all images: the median (Med.), mean, trimean (Tri.), best-25\% (B25\%), and worst-25\% (W25\%).
\vspace{0.25cm}
\par \noindent \textbf{Baselines and Training Settings.}
According to the classification of color constancy models in Section~\ref{sec2.2}, we selected several non-DNNCC models, including WP~\cite{land1977retinex}, SoG~\cite{finlayson2004shades}, GE~\cite{van2007edge}, Cheng et al.~\cite{cheng2015effective}, GI~\cite{Qian2019}, FFCC~\cite{barron2017fast}, RCC~\cite{li2023ranking}, as well as DNNCC models such as DS-Net~\cite{shi2016deep}, CNN~\cite{bianco2017single}, FC4~\cite{hu2017fc4}, Quasi-UCC~\cite{bianco2019quasi}, C4~\cite{yu2020cascading}, C5~\cite{afifi2021cross}, CLCC~\cite{lo2021clcc}, EIL-Net~\cite{cun2022learning}, SS (FC4)~\cite{buzzelli2024uncertainty}, ECF~\cite{buzzelli2024convolutional}, and GC\textsuperscript{3}~\cite{zhengguang2025gc3}  as comparison methods. Among them, the recently published or high-performance methods FC4, Quasi-UCC, C4, EIL-Net, ECF, and GC\textsuperscript{3} were selected as baseline models for our experiments.
The proposed BRE was integrated into all baseline models to assess its effectiveness in improving illuminant estimation performance.
Specifically, for Quasi-UCC, the brightness robustness enhancement excludes the contrastive loss component of BRE since brightness variation alters its ground truth. All DNNCC models were configured with a batch size of 16, while all other parameters were kept consistent with the official code. Training was conducted for 4000 epochs on the ColorChecker dataset and 2000 epochs on the Cube+ dataset. During the first 2000 epochs for the ColorChecker dataset and the first 1000 epochs for the Cube+ dataset as well as the brightness robustness dataset, the weights of adversarial loss $\lambda_{adv}$ and contrastive loss $\lambda_{ctr}$ (as defined in Eq.~\ref{eq12}) were set to 1 and 10, respectively. For the remaining epochs across all datasets, these weights were adjusted to 1 and 0.1. The random seed was fixed at 0, and all experiments were conducted on Nvidia GeForce RTX 3090 and A10 GPUs.

\begin{table}[t]
\renewcommand{\arraystretch}{1}
\centering
\scriptsize 
\setlength{\tabcolsep}{6pt}  
\begin{tabular}{l|l|ccccc}
\specialrule{1pt}{0pt}{0pt} 
\noalign{\smallskip} 
\multicolumn{2}{l|}{\textbf{Methods}}                                              & \textbf{Median} & \textbf{Mean} & \textbf{Trimean} & \textbf{Best 25\%} & \textbf{Worst 25\%} \\ \noalign{\smallskip}  
\hline \noalign{\smallskip} 
\multirow{6}{*}{\rotatebox{90}{\textbf{non-DNNCC}}} 
                                           & WP                     & 5.68            & 7.55          & 6.35             & 1.45               & 16.12               \\
                                           & GE                        & 4.44            & 5.13          & 4.62             & 2.11               & 9.26                \\
                                           & SoG                        & 2.94            & 3.67          & 3.03             & 0.99               & 7.75               \\
                                           & Cheng et al. 2014                 & 1.65            & 2.42          & 1.75             & 0.38               & 5.87                \\
                                           & RCC                           & 1.20            & 2.37          & 1.42             & 0.22               & 6.66                \\
                                           & FFCC                             & 1.10            & 2.00          & 1.40             & 0.30               & 5.10                \\ 
\noalign{\smallskip}  \hline \noalign{\vskip 1pt} \hline \noalign{\smallskip} 
\multirow{16}{*}{\rotatebox{90}{\textbf{DNNCC}}}    
                                           & CNN                              & 1.95            & 2.36          & -                & -                  & -                   \\
                                           & SS~(FC4)                               & 1.62            & 2.30          & 1.80             & -              & -                \\
                                           & C5                               & 1.61            & 2.36          & 0.44             & 5.60               & 1.74                \\
                                           & CLCC                               & 1.36            & 1.87          & 1.47             &  0.49              & 4.11                \\
                                           & DS-Net                           & 1.12            & 1.90          & 1.33             & 0.31               & 4.84                \\ 
\noalign{\smallskip}  
\cline{2-7} 
\noalign{\smallskip} 
  
                                           & \textbf{ECF}                              & 2.12               & 2.95             & 2.30                & 0.70                 & 6.57                   \\
                                           & \cellcolor{green!30}~+BRE                             & \cellcolor{green!30}1.94 (-8.74\%)                & \cellcolor{green!30}2.74 (-6.90\%)             & \cellcolor{green!30}2.10 (-8.86\%)                & \cellcolor{green!30}0.63 (-9.53\%)                  & \cellcolor{green!30}6.25 (-4.90\%)                   \\
                                           & \textbf{Quasi-UCC}                               & 1.48            & 2.24          & 1.66             & 0.41               & 5.30                \\
                                           & \cellcolor{green!30}~+BRE                            & \cellcolor{green!30}1.44 (-2.90\%)   & \cellcolor{green!30}2.11 (-5.52\%) & \cellcolor{green!30}1.59 (-3.89\%)    & \cellcolor{green!30}0.38 (-7.23\%)      & \cellcolor{green!30}4.95 (-6.49\%)       \\ 
                                           & \textbf{GC\textsuperscript{3}}                             & 1.43               & 2.33             & 1.62                & 0.40                  & 5.73                   \\
                                           & \cellcolor{green!30}~+BRE                             & \cellcolor{green!30}{1.36} (-4.80\%)               & \cellcolor{green!30}2.22 (-4.86\%)             & \cellcolor{green!30}1.54 (-5.28\%)                & \cellcolor{green!30}0.36 (-10.5\%)                  & \cellcolor{green!30}5.52 (-3.62\%)                   \\  
                                            & \textbf{FC4}            & 1.36               & 1.86             & 1.47                & 0.49                  & 4.06                   \\
                                           & \cellcolor{green!30}{~+BRE}                             & \cellcolor{green!30}1.27 (-7.02\%)               & \cellcolor{green!30}1.78 (-4.50\%)             & \cellcolor{green!30}1.37 (-6.62\%)                & \cellcolor{green!30}0.46 (-5.99\%)                  & \cellcolor{green!30}3.96 (-2.57\%)                   \\ 
                                           & \textbf{EIL-Net}                          & 1.15               & 1.63            & 1.24                & 0.38                  & 3.74                   \\
                                           & \cellcolor{green!30}~+BRE                             & \cellcolor{green!30}1.07 (-6.49\%)                & \cellcolor{green!30}1.60 (-2.39\%)             & \cellcolor{green!30}1.18 (-4.87\%)                & \cellcolor{green!30}0.36 (-6.62\%)                  & \cellcolor{green!30}3.74 (-0.00\%)                   \\ 
                                           & \textbf{C4}                             & 1.09               & 1.58             & 1.20                & 0.37                  & 3.61                   \\
                                           & \cellcolor{green!30}~+BRE                             & \cellcolor{green!30}{1.05} (-3.88\%)               & \cellcolor{green!30}1.52 (-3.55\%)             & \cellcolor{green!30}1.15 (-4.04\%)                & \cellcolor{green!30}0.35 (-4.38\%)                  & \cellcolor{green!30}3.49 (-3.40\%)                   \\

\noalign{\smallskip} 
\specialrule{1pt}{0pt}{0pt}
\end{tabular}
\caption{Comparative evaluation of color constancy methods with and without BRE strategy on ColorChecker datasets.} 
\label{tab2}
\end{table}
\subsection{Performance Evaluation of Illuminant Estimation} \label{sec5.1}
In this section, we evaluate how the proposed BRE influences the illuminant estimation performance of DNNCC models on standard color constancy datasets, with the goal of determining whether BRE can address performance limitations caused by insufficient brightness robustness.
\subsubsection{Quantitative Evaluation} \label{sec5.1.1}
Table~\ref{tab2} and Table~\ref{tab3} present the illuminant estimation performance metrics of various color constancy approaches—both DNN-based and non-DNN-based—on two datasets: ColorChecker (Table~\ref{tab2}) and Cube+ (Table~\ref{tab3}). The performance of the six selected baseline DNNCC models that incorporate BRE is shown in bold.

\par As shown in Table~\ref{tab2}, for the DNNCC models highlighted in bold on the ColorChecker dataset, incorporating the proposed BRE strategy yields substantial reductions in illumination estimation errors, with average decreases of 5.64\% and 4.62\% in the median and mean angular errors, respectively. 
Notably, for the current state-of-the-art C4 model, the BRE strategy achieves error reductions comparable to other baselines, setting a new benchmark for optimal performance. These experimental findings emphasize that brightness variation substantially constrains DNNCC models, whereas the proposed BRE strategy effectively mitigates this limitation by enhancing brightness robustness.

\begin{table}[t]
\renewcommand{\arraystretch}{1}
\centering
\scriptsize 
\setlength{\tabcolsep}{6pt}  
\begin{tabular}{l|l|ccccc}
\specialrule{1pt}{0pt}{0pt} 
\noalign{\smallskip} 
\multicolumn{2}{l|}{\textbf{Methods}}                                              & \textbf{Median} & \textbf{Mean} & \textbf{Trimean} & \textbf{Best 25\%} & \textbf{Worst 25\%} \\ \noalign{\smallskip}  
\hline \noalign{\smallskip} 
\multirow{6}{*}{\rotatebox{90}{\textbf{non-DNNCC}}} 
                                           & WP                     & 7.48            & 9.69          & 8.56             & 1.72               & 20.49               \\
                                           & GGW                           & 1.43            & 2.38          & 1.66             & 0.35               & 6.01                \\
                                           & SoG                        & 1.73            & 2.59          & 1.93             & 0.46               & 6.19               \\
                                           & GE                        & 1.59            & 2.50          & 1.78             & 0.48               & 6.08                \\
                                           & Cheng et al. 2014                 & 2.74            & 2.99          & 3.63             & -               & -                \\
                                           & GI                             & 1.39            & 1.64          & 2.39             & -               & -                \\ 
\noalign{\smallskip}  \hline \noalign{\vskip 1pt} \hline \noalign{\smallskip} 
\multirow{10}{*}{\rotatebox{90}{\textbf{DNNCC}}}    
                                           & \textbf{ECF}                              & 0.96               & 1.60             & 1.12                & 0.24                  & 4.02                   \\
                                           & \cellcolor{green!30}{~+BRE}                             & \cellcolor{green!30}0.82 (-14.44\%)               & \cellcolor{green!30}1.49 (-6.45\%)             & \cellcolor{green!30}0.98 (-11.88\%)                & \cellcolor{green!30}0.22 (-10.88\%)                  & \cellcolor{green!30}3.91 (-2.64\%)                   \\
                                           & \textbf{FC4}            & 1.17               & 1.55             & 1.26                & 0.42                  & 3.31                   \\
                                           & \cellcolor{green!30}{~+BRE}                             & \cellcolor{green!30}1.09 (-7.21\%)               & \cellcolor{green!30}1.50 (-3.32\%)             & \cellcolor{green!30}1.18 (-6.54\%)                & \cellcolor{green!30}0.39 (-8.52\%)                  & \cellcolor{green!30}3.31 (-0.05\%)                   \\ 

                                           & \textbf{Quasi-UCC}                               & 1.15            & 1.69          & 1.27             & 0.33               & 3.98                \\
                                           & \cellcolor{green!30}~+BRE                            & \cellcolor{green!30}1.09 (-5.48\%)   & \cellcolor{green!30}1.63 (-3.52\%) & \cellcolor{green!30}1.21 (-4.89\%)    & \cellcolor{green!30}0.31 (-6.75\%)      & \cellcolor{green!30}3.89 (-2.36\%)       \\ 
                                           & \textbf{EIL-Net}                          & 0.93               & 1.34             & 1.02                & 0.31                  & 3.09                   \\
                                           & \cellcolor{green!30}~+BRE                             & \cellcolor{green!30}0.91~(-2.43\%)              & \cellcolor{green!30}1.30~(-2.95\%)             & \cellcolor{green!30}0.99~(-2.43\%)                & \cellcolor{green!30}0.30~(-3.75\%)                  & \cellcolor{green!30}2.99~(-3.23\%)                   \\ 
                                           & \textbf{C4}                             & 0.97               & 1.40             & 1.06                & 0.33                  & 3.18                   \\
                                           & \cellcolor{green!30}~+BRE                             & \cellcolor{green!30}0.89 (-8.62\%)               & \cellcolor{green!30}1.31 (-6.34\%)             & \cellcolor{green!30}0.98 (-7.77\%)                & \cellcolor{green!30}0.29 (-10.28\%)                  & \cellcolor{green!30}3.05 (-4.27\%)                   \\ 
                                           & \textbf{GC\textsuperscript{3}}                            & 0.87               & 1.47             & 1.00                & 0.22                  & 3.74                   \\
                                           & \cellcolor{green!30}~+BRE                             & \cellcolor{green!30}0.73 (-15.82\%)               & \cellcolor{green!30}1.32 (-10.14\%)             & \cellcolor{green!30}0.86 (-13.74\%)                & \cellcolor{green!30}0.2 (-10.7\%)                  & \cellcolor{green!30}0.29 (-7.70\%)                   \\

\noalign{\smallskip} 
\specialrule{1pt}{0pt}{0pt}
\end{tabular}
\caption{Comparative evaluation of color constancy methods with and without BRE strategy on Cube+ datasets.} 
\label{tab3}
\end{table}

\par In Table~\ref{tab3}, the proposed BRE strategy also demonstrates measurable performance gains on the Cube+ dataset, underscoring the broad impact of brightness variation across different color constancy datasets and the substantial influence on existing DNNCC models.
Furthermore, we analyze the underlying mechanism through which the proposed BRE strategy operates from the perspective of DNNCC models. In part of the baseline models (FC4, C4, EIL-NET and GC\textsuperscript{3}), end-to-end architectures are employed to construct direct mappings from input images to illumination chromaticity for color constancy. Although these methods do not explicitly leverage brightness information, experimental results in Section~\ref{sec3} indicate their pronounced sensitivity to brightness variation, implying that brightness cues are inherently embedded during feature extraction. Consequently, BRE enhances the illumination estimation accuracy of these DNNCC models by suppressing the interference of brightness variation in the feature extraction process.
Similarly, ECF operates within a low-level statistical framework, performing illumination estimation by optimizing its parameters (e.g., Gaussian kernel functions, derivatives). Nevertheless, given that this optimisation is fundamentally reliant on RGB images, ECF remains unavoidably susceptible to brightness variation, which necessitates for performance enhancement through the proposed BRE strategy. Quasi-UCC, on the other hand, estimates illumination by indirectly identifying gray regions in brightness maps. Since it explicitly leverages brightness information, integrating BRE enables Quasi-UCC to extract more robust brightness features, thereby improving gray-region detection accuracy and boosting illumination estimation performance. 
This synergy is evident in Tables~\ref{tab2} and~\ref{tab3}, where BRE delivers significant performance gains for Quasi-UCC, underscoring the excellent compatibility between BRE and Quasi-UCC.

\begin{figure*}[t]
 \centering 
 \includegraphics[width=1\linewidth]{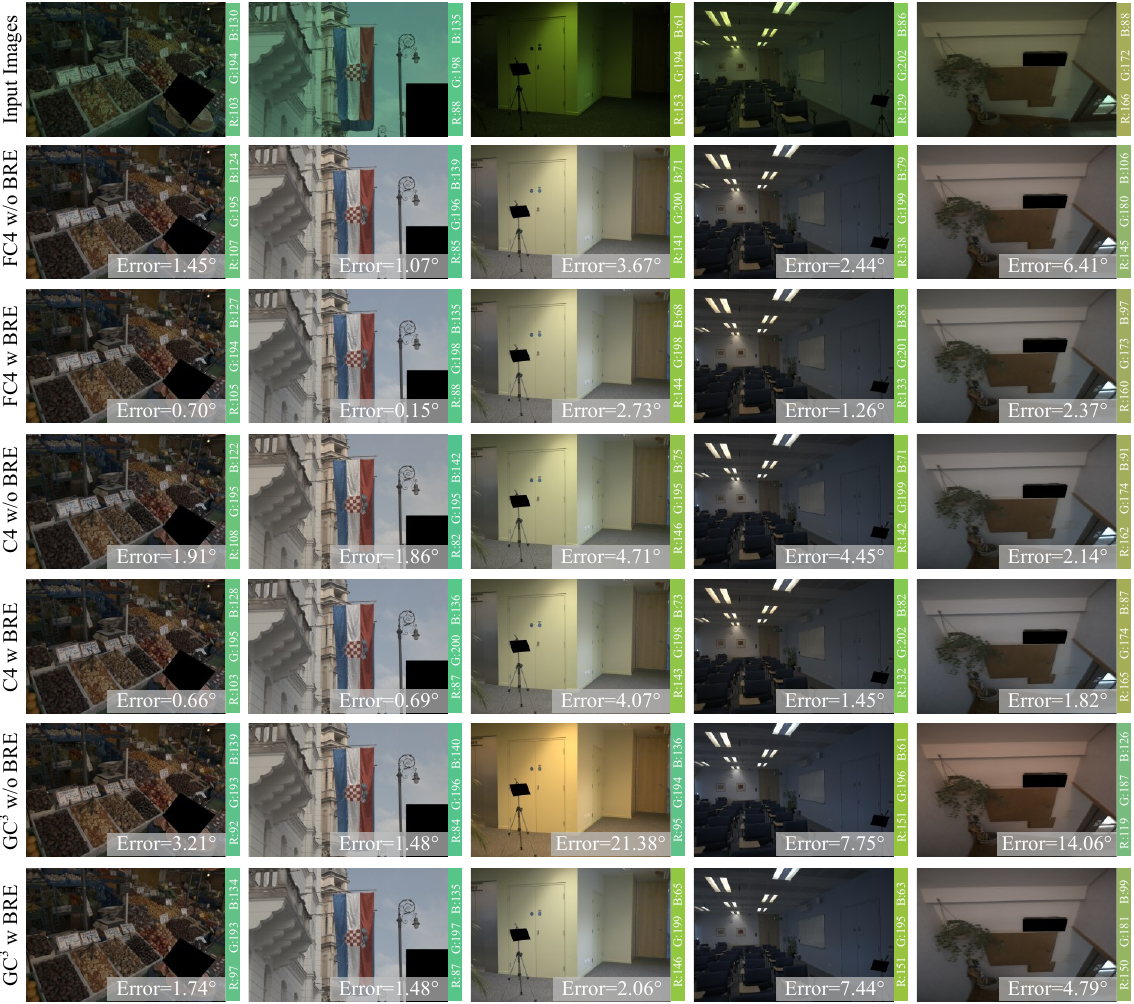} 
 \vspace{-0.7cm}
 \caption{Visual comparison of DNNCC models with angular errors in the lower right corner.}
 \label{fig5}
\end{figure*}

\subsubsection{Qualitative Evaluation} \label{sec5.1.2}
We conducted qualitative visual comparisons to evaluate the improvements achieved by integrating the proposed BRE into multiple DNNCC models.
As illustrated in Figure~\ref{fig5}, incorporating the BRE approach effectively improved illumination estimation accuracy, with corrected images exhibiting color consistency closer to the ground truth. Notably, BRE delivered greater performance gains for images with substantial brightness variations compared to other images. For instance, the images in columns 3-5 contain pronounced brightness disparities caused by spatial scene variations and artificial lighting. As discussed in Section~\ref{sec3}, such extreme brightness variations typically induce significant shifts in the DNNCC model’s extracted features, thereby degrading illumination estimation accuracy. By strengthening brightness robustness, the proposed BRE strategy ensures stable feature extraction under varying brightness conditions, effectively reducing estimation errors.

\subsection{Ablation Study and Analysis} \label{sec5.2}
\begin{table}[b]
\renewcommand{\arraystretch}{1}
\centering
\scriptsize 
\setlength{\tabcolsep}{5pt}  
\begin{tabular}{lll|ccccc|ccccc}
\specialrule{1pt}{0pt}{0pt} \noalign{\smallskip}
\multicolumn{3}{l|}{\multirow{2}{*}{\begin{tabular}[c]{@{}l@{}}\textbf{Ablation} \\ \textbf{Study}\end{tabular}}} & \multicolumn{5}{c|}{\textbf{ColorChecker}}                                                                      & \multicolumn{5}{c}{\textbf{Cube+}}                                                                        \\ \noalign{\smallskip} \cline{4-13} \noalign{\smallskip}
\multicolumn{3}{l|}{}                                                                           & \textbf{Med.}  & \textbf{Mean}  & \multicolumn{1}{l}{\textbf{Tri.}} & \multicolumn{1}{l}{\textbf{B25\%}} & \multicolumn{1}{l|}{\textbf{W25\%}} & \textbf{Med.}  & \textbf{Mean}  & \multicolumn{1}{l}{\textbf{Tri.}} & \multicolumn{1}{l}{\textbf{B25\%}} & \multicolumn{1}{l}{\textbf{W25\%}} \\ \noalign{\smallskip} \hline \noalign{\smallskip}
\multicolumn{3}{l|}{\textbf{Baseline}}                                                                   & 1.09 & 1.58 & 1.20                    & 0.37                     & 3.61                     & 0.97 & 1.40 & 1.06                    & 0.33                     & 3.18                     \\
\multicolumn{3}{l|}{\textbf{w/o ABA}}                                                                    & 1.13 & 1.58 & 1.22                    & 0.39                     & 3.56                     & 0.99 & 1.40 & 1.08                    & 0.34                     & 3.17                     \\
\multicolumn{3}{l|}{\textbf{w/o BCL}}                                                                    & \textcolor{blue}{1.07} & \textcolor{blue}{1.54} & \textcolor{blue}{1.16}                    & \textcolor{blue}{0.36}                     & \textcolor{blue}{3.52}                     & \textcolor{blue}{0.93} & \textcolor{blue}{1.35} & \textcolor{blue}{1.02}                    & \textcolor{blue}{0.30}                     & \textcolor{blue}{3.11}                     \\
\multicolumn{3}{l|}{\textbf{Full}}                                                                       & \textcolor{red}{1.05} & \textcolor{red}{1.52} & \textcolor{red}{1.15}                    & \textcolor{red}{0.35}                     & \textcolor{red}{3.49}                     & \textcolor{red}{0.89} & \textcolor{red}{1.31} & \textcolor{red}{0.98}                    & \textcolor{red}{0.29}                     & \textcolor{red}{3.05}                     \\ \noalign{\smallskip} \specialrule{1pt}{0pt}{0pt}
\end{tabular}
\caption{Comparison of ablation studies with C4 on the ColorChecker and Cube+ dataset. Optimal and suboptimal results are highlighted in \textcolor{red}{red} and \textcolor{blue}{blue}, respectively.}
\label{tab4}
\end{table}
In this section, we initially investigate two key components of BRE—Adversarial Brightness Augmentation (ABA) and Brightness Contrast Loss (BCL) through ablation experiments. ABA is designed to identify high-risk brightness images for data augmentation, while BCL enhances brightness robustness by minimizing the feature distance between original and brightness-augmented images. The experiments are organized as follows:

\begin{itemize}
   \item 
   Without Adversarial Brightness Augmentation (w/o ABA): the ABA is replaced by a random brightness transformation.
   \item 
   Without Brightness Contrast Loss (w/o BCL): The model is trained exclusively using adversarially augmented images.
\end{itemize}
\par As shown in Table~\ref{tab4}, random brightness transformation (w/o ABA) slightly degrades model performance compared to the baseline. In contrast, adversarial brightness transformation demonstrates clear performance improvements, reducing the mean angular error from 1.09 to 1.05 on the ColorChecker dataset and from 0.97 to 0.89 on the Cube+ dataset. This superior performance is attributed to its ability to optimize brightness parameters based on the model's responses, thereby discovering model-specific high-risk brightness variation and enabling targeted refinements. Conversely, random brightness transformation may interfere with the model's learning due to indiscriminate augmentation.
Furthermore, omitting the brightness contrast loss from BRE results in performance degradation across all evaluation metrics. As illustrated in Table~\ref{tab4}, the C4 model exhibits increased median angular errors of 1.9\% and 4.5\% on the ColorChecker and Cube+ datasets respectively when this component is removed. This suggests that brightness contrast loss effectively suppresses feature shifts under brightness variation by imposing feature consistency constraints, further enhancing the model’s brightness robustness. Notably, even without brightness contrast loss, the proposed BRE strategy still surpasses the baseline by a considerable margin, underscores the effectiveness of adversarial brightness transformation in preserving robust brightness features. However, the complete BRE strategy achieves superior performance, highlighting the synergistic interplay between brightness contrast loss and adversarial brightness transformation.

\begin{figure*}[t]
 \centering 
 \includegraphics[width=1\linewidth]{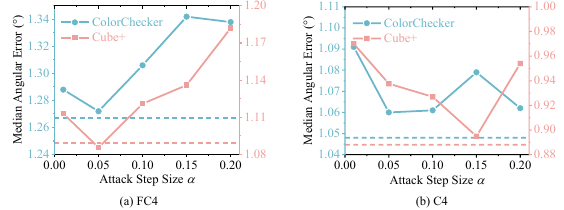} 
 \vspace{-0.7cm}
 \caption{Evaluation of the impact of attack step sizes $\alpha$ on FC4~(a) and C4~(b), where dashed lines indicate the performance with proposed adaptive attack step sizes.}
 \label{fig6}
\end{figure*}

\par Subsequently, we evaluate the effectiveness of our proposed adaptive step size for adversarial brightness augmentation. Specifically, we compare its performance with that of several fixed step sizes (0.01, 0.05, 0.1, 0.15, 0.2). Figure~\ref{fig6} illustrates the median angular error curves of the DNNCC models~(FC4 and C4) for each step size, with dashed lines representing the results obtained using the adaptive step size strategy.
As illustrated in Figure~\ref{fig6}, the optimal step size varies across different models and datasets. For instance, the optimal attack step size for C4 on the ColorChecker dataset is 0.05, while on the Cube+ dataset it is 0.15. Moreover, even within the same dataset, the optimal step size can differ between DNNCC models. For example, in the Cube+ dataset, the optimal attack step size for FC4 is 0.05, whereas for C4, it is 0.15.
These findings suggest that a single fixed step size may not be suitable for all DNNCC models and datasets, potentially limiting improvements in model performance.
Conversely, with the proposed adaptive step size strategy, both the FC4 and C4 models achieve optimal or near-optimal performance on the ColorChecker and Cube+ datasets. Such results underscore the robust generalization capability of the adaptive step-size strategy in improving the illumination robustness of DNNCC models.

\subsection{Evaluation of Brightness Robustness Enhancement} \label{sec5.4}
\begin{table}[t]
\renewcommand{\arraystretch}{1}
\centering
\scriptsize 
\setlength{\tabcolsep}{8pt}  
\begin{tabular}{l|ccccc}
\specialrule{1pt}{0pt}{0pt} 
\noalign{\smallskip} 
\textbf{Methods}                                              & \textbf{Median} & \textbf{Mean} & \textbf{Trimean} & \textbf{Best 25\%} & \textbf{Worst 25\%} \\ \noalign{\smallskip}  
\hline \noalign{\smallskip} 
\textbf{ECF} & 1.54               & 2.74             & 1.83                & 0.44                  & 7.05                   \\
\cellcolor{green!30}~+BRE                & \cellcolor{green!30}1.10 (-28.45\%)               & \cellcolor{green!30}1.49 (-45.64\%)             & \cellcolor{green!30}1.20 (-34.59\%)                & \cellcolor{green!30}0.41 (-5.69\%)                  & \cellcolor{green!30}3.18 (-54.82\%)                   \\ 
\textbf{GC\textsuperscript{3}}        & 1.32               & 2.27             & 1.44                & 0.40                  & 5.87                   \\
\cellcolor{green!30}~+BRE                & \cellcolor{green!30}1.05(-20.40\%)               & \cellcolor{green!30}2.00 (-11.52\%)             & \cellcolor{green!30}1.18 (-17.54\%)                & \cellcolor{green!30}0.34 (-15.42\%)                  & \cellcolor{green!30}4.98 (-15.12\%)                   \\ 
\textbf{Quasi-UCC} & 0.96            & 2.96         & 1.40             & 0.18               & 8.89                \\
\cellcolor{green!30}~+BRE                & \cellcolor{green!30}0.76 (-21.39\%)   & \cellcolor{green!30}2.63 (-11.10\%) & \cellcolor{green!30}1.27 (-8.89\%)    & \cellcolor{green!30}0.17 (-6.57\%)      & \cellcolor{green!30}8.11 (-8.80\%)       \\ 

\textbf{C4}       & 0.68               & 1.16             & 0.74                & 0.17                  & 2.95                   \\
\cellcolor{green!30}~+BRE                & \cellcolor{green!30}0.45 (-33.89\%)               & \cellcolor{green!30}0.71 (-38.82\%)             & \cellcolor{green!30}0.50 (-31.86\%)                & \cellcolor{green!30}0.12 (-27.16\%)                  & \cellcolor{green!30}1.74 (-41.01\%)                   \\ 
\textbf{FC4}            & 0.44               & 0.81             & 0.51                & 0.12                  & 2.11                   \\
\cellcolor{green!30}{~+BRE}              &  \cellcolor{green!30} 0.35 (-20.05\%)               & \cellcolor{green!30}0.59 (-27.34\%)             & \cellcolor{green!30}0.40 (-22.82\%)                & \cellcolor{green!30}0.11 (-4.18\%)                  & \cellcolor{green!30}1.46 (-30.77\%)                   \\
\textbf{EIL-Net}  & 0.45               & 0.99             & 0.55                & 0.10                  & 2.78                   \\
\cellcolor{green!30}~+BRE                & \cellcolor{green!30}0.22 (-51.66\%)               & \cellcolor{green!30}0.60 (-39.68\%)             & \cellcolor{green!30}0.28 (-48.57\%)                & \cellcolor{green!30}0.09 (-15.24\%)                  & \cellcolor{green!30}1.74 (-37.29\%)                   \\ 
\noalign{\smallskip} 
\specialrule{1pt}{0pt}{0pt}
\end{tabular}
\caption{Brightness robustness comparison of color constancy methods with and without BRE
}
\label{tab5}

\end{table}

\begin{figure}[t]
  \includegraphics[width=1\linewidth]{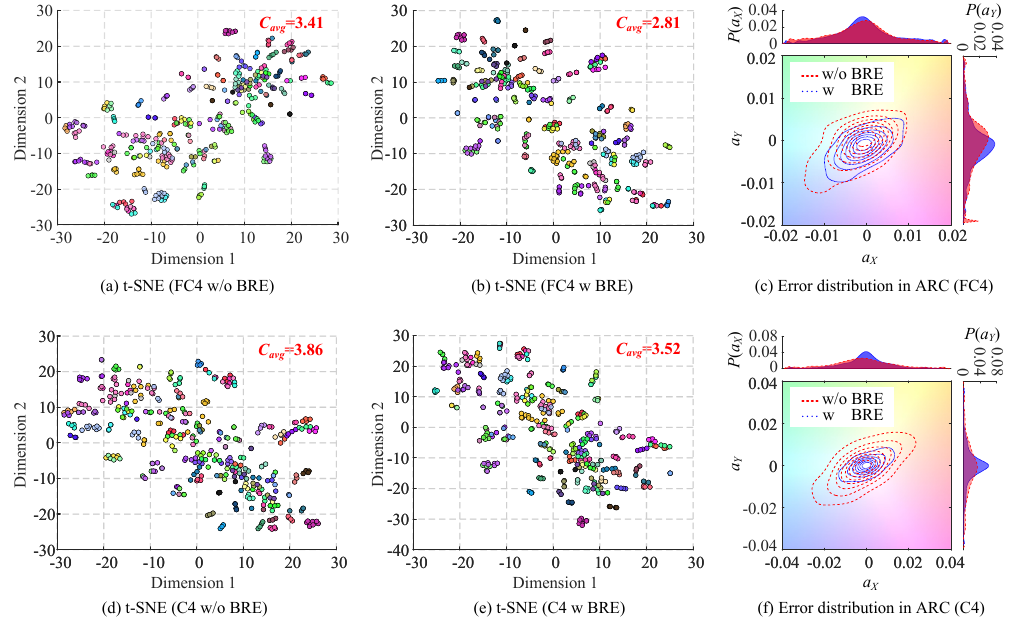}
  \vspace{-0.7cm}
  \centering
  \caption{Analysis of illumination robustness through feature and error distributions of DNNCC models. t-SNE feature visualizations~\cite{vandermaaten2008visualizing} for (a, b) FC4 and (d, e) C4, where feature points are color-coded according to their labels. A lower intra-class scatter, $C_{avg}$, in the top-right corner indicates more consistent predictions for the same labels, suggesting stronger brightness robustness. (c, f) Error distributions in the ARC coordinate system~\cite{buzzelli2020arc} coordinates illustrate FC4 and C4 errors with and without the BRE strategy, where distributions closer to the center reflect higher prediction accuracy. $P(\alpha_X)$ and $P(\alpha_Y)$ denote the probability distributions of $\alpha_X$ and $\alpha_Y$ along the respective axes.}
  \label{fig7}
\end{figure}
This section evaluates the effectiveness of the proposed BRE strategy in enhancing DNNCC models' brightness robustness. We compare the performance of DNNCC models with and without BRE on our constructed brightness robustness dataset. The experiments follow the same train-test split scheme as described in Section~\ref{sec3}.
As illustrated in Table~\ref{fig5}, the incorporation of BRE substantially improves the illumination estimation accuracy of all DNNCC models, with average reductions of 29.31\% and 29.02\% in median and mean angular errors, respectively. These results demonstrate the effectiveness of the proposed BRE strategy in enhancing model robustness under brightness variation.
\par To provide deeper insights into the brightness robustness improvements conferred by BRE, we visualize the feature distributions of C4 and FC4 using t-SNE in Figure~\ref{fig7}(a,b,d,e), along with their illuminant error distributions in Angle-Retaining Chromaticity (ARC) cartesian coordinates (Figure~\ref{fig7}(c, f)), comparing the conditions with and without BRE. Figure~\ref{fig7}(a), (d) illustrate that DNNCC models without BRE exhibit scattered feature distributions, where images with identical ground truth labels struggle to form coherent clusters in feature space. This scattering reflects the instability of the DNNCC model's feature extraction when processing images with identical ground truth but varying brightness levels in the absence of BRE. In contrast, models incorporating BRE demonstrate markedly improved feature distributions. As shown in Figure~\ref{fig7}(b) and (e), images with identical ground truth labels form distinct, compact clusters in the feature space. This observation is further supported by the lower intra-class scatter ($C_{avg}$) shown in the upper right corners of the subplots, where BRE reduces $C_{avg}$ of FC4 and C4 from 3.41 to 2.81 and 3.86 to 3.52, respectively, indicating that the proposed BRE effectively enhances feature consistency under brightness variation.
Examining the error distributions shown in (c) and (f), the DNNCC models without BRE exhibit a considerably larger error range in the ARC coordinate system. Notably, in the absence of BRE, the C4 model also displays a distinct positive skew on the $a_X$ axis, indicating a systematic prediction bias under varying brightness conditions. In contrast, after incorporating BRE, the error distributions of DNNCC models demonstrate significantly reduced ranges and markedly decreased skewness, further validating BRE's effectiveness in enhancing model robustness against brightness variation.

\section{Conclusion}\label{sec6}
In this paper, we identify a potential challenge associated with brightness robustness in color constancy. By constructing a specialized dataset focused on brightness robustness, we conduct a pioneering investigation into the performance of DNNCC models under varying brightness conditions, uncovering the limitations caused by such variations.
To tackle this challenge, we propose a Brightness Robustness Enhancement (BRE) strategy, which identifies high-risk brightness variation and incorporates brightness-robustness-aware optimization to improve DNNCC models' performance.
Evaluation across two public color constancy datasets shows that the proposed BRE significantly enhances the models’ performance and brightness robustness, highlighting its potential for developing more robust and efficient DNNCC models. We hope this research raises awareness of the potential robustness risks associated with DNNCC models. Moving forward, we aim to develop more realistic brightness variation simulations to further enhance DNNCC models' performance.

\section*{Funding}
This work was supported in part by the National Natural Science Foundation of China under Grants 62394330 and 62072126, in part by the Guangdong Basic and Applied Basic Research Foundation under Grant 2024A1515012064, in part by the Fundamental Research Projects Jointly Funded by Guangzhou Council and Municipal Universities under Grant 2024A03J0394, in part by the Fundamental Research Projects Funded by Liwan Institute under Grant LWYJ202418.

\bibliography{main}
\end{document}